\documentclass[sigconf]{acmart}

\renewcommand\footnotetextcopyrightpermission[1]{}

\acmConference{}{}{}
\acmBooktitle{}
\AtBeginDocument{%
  }

\usepackage{graphicx}
\usepackage{adjustbox}
\usepackage{subcaption}
\usepackage{makecell}
\usepackage{tcolorbox}
\tcbuselibrary{breakable}
\usepackage{mdframed}
\usepackage{xcolor}
\usepackage{enumitem}
\newtcolorbox{compactprompt}{
    colback=gray!10,      
    colframe=gray!50,     
    boxrule=0.5pt,        
    arc=0pt,              
    boxsep=1pt,           
    left=3pt, right=3pt,  
    top=2pt, bottom=2pt,  
    fontupper=\ttfamily\footnotesize, 
    breakable             
}

\begin{document}

\title{Toward Expert Investment Teams: \\A Multi-Agent LLM System with Fine-Grained Trading Tasks}

\author{Kunihiro Miyazaki}
\orcid{0000-0001-7800-8019}
\affiliation{%
  \institution{Japan Digital Design, Inc.}
  \city{Tokyo}
  \country{Japan}
}
\email{kunihirom@acm.org}

\author{Takanobu Kawahara}
\affiliation{%
  \institution{Japan Digital Design, Inc.}
  \city{Tokyo}
  \country{Japan}
}
\email{takanobu.kawahara@japan-d2.com}

\author{Stephen Roberts}
\orcid{0000-0002-9305-9268}
\affiliation{%
  \institution{Department of Engineering Science \\ University of Oxford}
  \country{United Kingdom}
}
\email{sjrob@robots.ox.ac.uk}

\author{Stefan Zohren}
\orcid{0000-0002-3392-0394}
\affiliation{%
 \institution{Department of Engineering Science \\ University of Oxford}
  \country{United Kingdom}
}
\email{stefan.zohren@eng.ox.ac.uk}

\renewcommand{\shortauthors}{Miyazaki et al.}

\begin{abstract}
The advancement of large language models (LLMs) has accelerated the development of autonomous financial trading systems.
While mainstream approaches deploy multi-agent systems mimicking analyst and manager roles, they often rely on abstract instructions that overlook the intricacies of real-world workflows, which can lead to degraded inference performance and less transparent decision-making.
Therefore, we propose a multi-agent LLM trading framework that explicitly decomposes investment analysis into fine-grained tasks, rather than providing coarse-grained instructions.
We evaluate the proposed framework using Japanese stock data, including prices, financial statements, news, and macro information, under a leakage-controlled backtesting setting.
Experimental results show that fine-grained task decomposition significantly improves risk-adjusted returns compared to conventional coarse-grained designs.
Crucially, further analysis of intermediate agent outputs suggests that alignment between analytical outputs and downstream decision preferences is a critical driver of system performance.
Moreover, we conduct standard portfolio optimization, exploiting low correlation with the stock index and the variance of each system's output. This approach achieves superior performance.
These findings contribute to the design of agent structure and task configuration when applying LLM agents to trading systems in practical settings.  \looseness=-1
\end{abstract}



\maketitle

\section{Introduction}
The rapid advancement of Large Language Models (LLMs) has led to high expectations that they function as an autonomous workforce, like human employees, in practical domains~\cite{WhartonP46:online,AllofMyE96:online}.
Consequently, many companies have initiated the internal and external implementation of ``AI agents''~\cite{TheState62:online,brynjolfsson2025generative,peng2023impact}. 
In the financial industry, the development of AI agents has begun across various applications~\cite{nie2024survey,dong2025large}, with expectations to generate profits in both investment and trading contexts through the utilization of agent-based autonomy and intelligence~\cite{lopez2025can,xiao2024tradingagents}.
One of the most common configurations is a multi-agent trading system, in which multiple LLM agents are assigned specific roles such as fundamental analysis, news parsing, and risk management~\cite{xiao2024tradingagents,zhao2025alphaagents,yu2024fincon}.  \looseness=-1

However, considering practical applications, concerns exist that current LLM trading systems are constructed with a simplified task design, given the complexity of investment analyst tasks. 
Most existing studies adopt coarse-grained task settings, primarily assigning roles and high-level objectives to agents. 
For instance, a fundamental agent tends to be simply instructed to ``analyze financial statements (e.g., 10-K)'', leaving unexplored the fine-grained task design for both qualitative and quantitative analyses typically performed in real-world situations~\cite{xiao2024tradingagents,zhao2025alphaagents,yu2024fincon}.  \looseness=-1

Providing coarse-grained instructions to LLMs for complex tasks presents two major challenges. 
The first is performance degradation. 
It has been reported that overly vague instructions can reduce the output quality of LLMs~\cite{kim2024aligning,zhou2022least}, and when tasks are too complex, LLMs have been observed to occasionally cease reasoning midway or abandon reasoning entirely~\cite{shojaee2025illusion,wen2024language}. 
The second is the lack of interpretability. 
When LLMs are given ambiguous instructions, typically only the final output is visualized, making it impossible to interpret the intermediate reasoning process~\cite{zhao2024explainability,palikhe2025towards}. 
In such cases, practical deployment becomes difficult, especially in asset management practices where significant capital is at stake~\cite{jadhav2025large}. \looseness=-1

To address these issues, this study constructs a multi-agent LLM trading system that assigns detailed, concrete investment decision-making tasks to trading agents, based on real-world practices of investment analysts. 
In contexts beyond finance, it has been reported that providing expert processes to LLM agent systems is effective~\cite{hong2023metagpt}.
Furthermore, separating \textit{task planning} from \textit{execution} is reported to be effective in LLM agent systems~\cite{wang2023plan,qiu2025blueprint}. 
It is also noted that performance improves when domain experts design the tasks~\cite{li2025decomposed}. 
Inspired by these studies, we anticipate that tracing the complex workflows of real-world investment analysts will enhance the performance of investment agents. 
Furthermore, responding to concrete tasks is expected to improve the agents' output explainability~\cite{yao2022react,ku2025multi,he2025unleashing}, which is crucial for industrial applications~\cite{jadhav2025large}. \looseness=-1

In the experiment, we conduct backtesting to evaluate whether our proposed fine-grained task configuration leads to performance improvements.
The evaluation uses Japanese equity market data, including stock prices, financial statements, news articles, and macroeconomic information. 
To prevent data leakage and account for the LLM model's knowledge cutoff, we set the backtesting period from September 2023 to November 2025.
Beyond portfolio-level performance, we also evaluate intermediate textual outputs to analyze how task decomposition affects reasoning behavior and interpretability. 
Finally, to demonstrate real-world applicability, we verify the strategy's effectiveness through portfolio optimization benchmarked against market indices.

This study makes the following contributions.
\textbf{Impact of Task Granularity:} We focus on task design for LLM-based trading agents, which has been largely overlooked in prior works, and demonstrate through controlled experiments that fine-grained task decomposition improves agent performance.
\textbf{Agent Ablation Analysis:} We conduct a comprehensive ablation study by systematically removing and replacing individual agents, providing new insights into the functional roles of agents in multi-agent trading systems.
\textbf{Real-world Evaluation:} We emphasize real-world applicability by adopting realistic problem settings (e.g., agent roles and team architecture) and evaluating not only portfolio-level performance but also intermediate textual outputs and portfolio optimization results benchmarked against market indices.
\textbf{Reproducibility:} To support reproducibility and future research, we release the implementation codes with prompts upon acceptance.

\section{Related Work}
\subsection{Multi-Agent Trading Systems with LLMs}
In the emergence of LLM trading systems, whereas early studies primarily adopted single-agent architectures~\cite{lopez2025can}, more recent work has shifted toward multi-agent trading systems that more closely resemble real investment teams~\cite{saha2025large}.
In such systems, multiple agents are assigned complementary roles and collaborate to process heterogeneous financial information~\cite[e.g.][]{xiao2024tradingagents,yu2024fincon,xiong2025quantagent,zhao2025alphaagents,kundu2025multi}. \looseness=-1

The focus of existing research on multi-agent LLM trading can broadly be grouped into two categories~\cite{saha2025large}.
The first category involves the efforts concerning their organizational structure and roles, such as agents' arrangements, interactions between agents, and role diversification, often inspired by real financial institutions.
These typically adopt a manager–analyst architecture, where a manager coordinates several specialized analyst agents that collect, filter, and synthesize information from various sources~\cite{yu2024fincon}.
Analyst roles span a wide range of functions, including fundamental analysis~\cite{zhao2025alphaagents}, technical analysis~\cite{xiong2025quantagent}, news sentiment extraction~\cite{xiao2024tradingagents}, and risk management~\cite{li2025hedgeagents,yu2025finmem}.
The second category focuses on reinforcement learning-based approaches, which aim at improving decision policies through iterative feedback.
Notable directions include Reflection mechanisms, which incorporate realized trading outcomes into subsequent reasoning, and Layered Memory architectures that regulate the temporal scope of accessible information~\cite{tian2025tradinggroup,zhang2024multimodal,yu2024fincon}. \looseness=-1

Our research builds on the first research focus, namely, structure and roles.
While prior studies carefully design both the structure and roles of agents, the prompts that specify how each role should operate have not been fully explored---they are often defined at a relatively coarse level, without explicitly aligning them with real investment tasks.
This is presumably because the field of multi-agent systems for LLM-based trading is nascent; research has primarily focused on achieving end-to-end trading completion, with little light shed on the granular details of task execution.
In contrast, we aim to formulate role prompts in a more fine-grained and realistic manner that deliberately imitates the division of labor in real investment organizations, to improve controllability and interpretability of trading performance.  
Furthermore, despite the growing interest in agent-based LLM trading systems, we believe that agent-level ablation studies remain relatively underexplored; thus, by conducting a systematic analysis, we provide important practical insights. \looseness=-1

\subsection{Prompt Designs and Expert Task Settings}
Recent LLM research explores whether explicitly planning and decomposing problems (rather than providing simple and ambiguous instructions) can improve performance on complex tasks.
Frameworks such as MetaGPT~\cite{hong2023metagpt} and Agent-S~\cite{agashe2024agent} demonstrate that encoding Standard Operating Procedures (SOPs) into multi-agent systems can reduce errors and enhance output quality, particularly in software-engineering tasks.
Related approaches, including the ``plan-and-execute''~\cite{wang2023plan} and ``blueprint-first''~\cite{qiu2025blueprint} paradigms, further suggest that fixing workflow structures—rather than allowing LLMs to autonomously determine task sequences—helps stabilize long-context reasoning.
Taken together, these results indicate that embedding expert knowledge into prompts can improve reliability and performance, consistent with findings reported in~\citet{li2025decomposed}. \looseness=-1

In contrast, the financial domain is only beginning to explore how such ``expert processes'' can be formalized for LLMs.
Existing approaches often rely on agents inspired by real or virtual investor personas~\cite{yu2025finmem,chen2025finhear} or on systems that allow humans to intervene during reasoning~\cite{wang2025alpha}.
However, these systems typically do not treat expert workflows themselves as explicit structural components. \looseness=-1

The work most closely related to ours is Financial Chain-of-Thought (CoT) prompting introduced in FinRobot~\cite{yang2024finrobot,kim2024financial}.
By pre-structuring financial analysis into predefined sections, FinRobot encourages domain-specific reasoning when generating analysis reports.
Our approach differs from the prior work in several respects.
First and foremost, we formalize expert workflows as fixed analysis protocols, rather than relying on generic CoT-style prompting.
Second, rather than focusing on report generation, we aim to operationalize trading decisions.
Third, we extend beyond firm-level analysis by incorporating other factors such as macroeconomic information and sector information. \looseness=-1

\section{Problem Setting}
The primary aim of this study is to investigate whether providing fine-grained task specifications to LLM agents in automated trading contributes to improved operational performance. 
To validate the effectiveness of the multi-agent system with the proposed setting, we conduct evaluations under constraints that mimic the investment practices of institutional investors. 
Specifically, we construct and backtest a monthly rebalancing portfolio based on a long-short strategy targeting large-cap stocks in Japanese equity markets. \looseness=-1

\subsection{Backtesting and Model Setup}
\textbf{Investment Universe:} 
We use the TOPIX 100 constituents, representing stocks with the highest market capitalization in Japan.

\noindent \textbf{Portfolio Construction:} 
To eliminate market-wide volatility risk and isolate stock selection capability, we adopt a market-neutral strategy. 
Specifically, the portfolio holds an equal number of stocks for both long (buy) and short (sell) positions with equal weighting.

\noindent \textbf{Rebalancing Frequency:} 
We conduct portfolio rebalancing at the opening of the first business day of each month.

\noindent \textbf{Test Period:} The evaluation covers the period from September 2023 to November 2025, totaling 27 months.

\noindent \textbf{Model Selection and Look-ahead Bias Mitigation:} 
We employ state-of-the-art LLMs, specifically the GPT-4o~\cite{hurst2024gpt} with the knowledge cutoff date in August 2023, as an inference model. 
Previous research warns that LLMs may ``memorize'' historical financial time-series data present in their training corpora, posing a risk of look-ahead bias that artificially inflates backtesting results~\cite{lopez2025memorization}. 
To guarantee the validity of our backtest, it is crucial to strictly separate the agent's knowledge from future events~\cite{hwang2025decision,drinkall2024time}. 
In this experiment, we enforce strict temporal ordering by feeding the agents only text and market data publicly available up to the specific decision point and by conducting the backtesting period after the LLM's knowledge cutoff date, thereby preventing information leakage. 
For the model setting, we use the default, such that the temperature is set to unity. 
While a temperature of zero might seem preferable for reducing variation, it does not fully eliminate stochasticity~\cite{ouyang2025empirical,atil2024non}. 
Since we aggregate multiple outputs using the median in our experiment, variability is already mitigated. 
Also, from an ensemble perspective, a temperature of unity can be preferable as it preserves useful output diversity~\cite{wang2022self,chen2021evaluating}.
As for the inference time, we note that since we conduct a monthly rebalancing, real-time processing is unnecessary.

\subsection{Evaluation Metrics}
We assess the performance of the agent-generated portfolios using both quantitative and qualitative metrics.

\noindent \textbf{Quantitative Metrics:} 
We use the standard measure of risk-adjusted return, namely the Sharpe ratio.
We calculate it as the mean of monthly portfolio returns divided by their standard deviation.

\noindent \textbf{Qualitative Metrics:} 
We analyze the scores and reasoning texts generated by the analyst agents to assess how these agents communicate with the manager agents. 
\looseness=-1

\section{Multi-Agent Framework Configuration}
Prior to detailing the fine-grained tasks assigned to each agent, this section outlines our baseline configuration of the LLM investment team. 
While multi-agent systems generally allow for flexible composition of agents~\cite[e.g.,][]{yu2024fincon,zhao2025alphaagents,xiao2024tradingagents,tian2025tradinggroup,xiong2025quantagent}, from a practical industry perspective, we aim to assemble the realistic components emulating the operational workflows of professional institutional investors. 

\subsection{Hierarchical Decision-Making Process}
\begin{figure*}[!ht]
    \centering
    \includegraphics[width=0.9\linewidth]{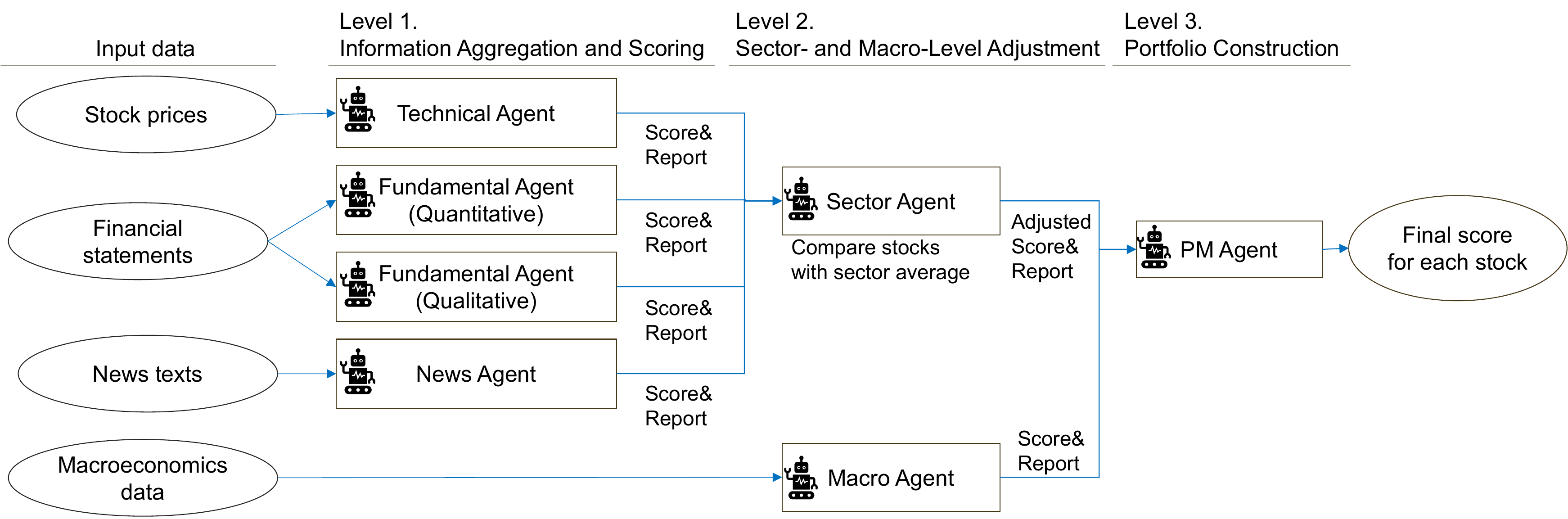}
    \caption{Overview of our multi-agent LLM trading system (see main text for details).}
    \label{fig:framework}
\end{figure*}

The system adopts a bottom-up manager-analyst framework for decision-making, where information is progressively abstracted and aggregated through a multi-level hierarchical structure from analysts to the portfolio manager.
Figure~\ref{fig:framework} depicts an overview of our trading system composition.

\noindent \textbf{Level 1. Information Aggregation and Scoring (Analyst Agents):} 
Four types of specialist agents---Quantitative, Qualitative, News, and Technical---analyze each stock in the investment universe. 
The Quantitative and Technical Agents assign attractive scores $S \in [0, 100]$ and generate a textual rationale supporting their evaluation based on their specific domain expertise.
The Qualitative and News Agents generate the supplemental information (scores and texts).
The output scores and reports are integrated and submitted to the Sector Agent.

\noindent \textbf{Level 2. Sector-Level Adjustment (Sector Agent):} 
The Sector Agent aggregates scores and reports from subordinate analysts and adjusts them based on sector-specific benchmarks. 
Specifically, a stock's quantitative valuation score is re-evaluated against sector averages.
The adjusted scores and reports are submitted to the Portfolio Manager (PM) agent.

\noindent \textbf{Level 2. Macro-Environmental Assessment (Macro Agent):} 
Independently, the Macro Agent analyzes broader economic indicators, such as interest rate trends, business cycles, and foreign exchange dynamics. 
It evaluates the current market regime and submits scores and texts to the PM Agent.

\noindent \textbf{Level 3. Final Portfolio Construction (PM Agent):} 
The PM Agent synthesizes the scores and reports from the Sector Agent and the Macro Agent to determine final scores for all stocks in the TOPIX 100. 
Then, we select the same number of stocks for long positions with the highest scores and the short positions with the lowest scores to construct a portfolio. 

\subsection{Data Sources}
The agents access a combination of structured and unstructured data to inform their decisions.
Considering reproducibility, we adhere to open data as much as possible.

\noindent \textbf{Stock Price Data:} 
We use daily prices of stocks in TOPIX 100 obtained from Yahoo Finance~\cite{ranarous38:online}. 
We use close prices to calculate the metrics for investment decisions, and use open prices for execution and performance evaluation.
To clarify, we conduct rebalancing after the market close of the last business day of a month and execute it at the open of the first business day of the following month.

\noindent \textbf{Financial Statements Data:} 
We utilize quarterly, semi-annual, and annual securities reports of companies. 
We retrieve these documents through the EDINET API~\cite{EDINET88:online} managed by the Financial Services Agency (FSA) of Japan.
We use both numerical financial statements and qualitative textual information. 

\noindent \textbf{News Data:} 
To capture sentiment and events (e.g., scandals, product launches) related to companies, we aggregate headlines and freely accessible article previews from major financial news outlets, including \textit{Nikkei}, \textit{Reuters}, and \textit{Bloomberg} (all Japanese editions) sourced from Ceek.jp News~\cite{CeekjpNe13:online}, a comprehensive aggregator of virtually all Japanese news media articles.

\noindent \textbf{Macroeconomic Data:} 
To gauge the macroeconomic environment, we compile a comprehensive dataset using the FRED~\cite{StLouisF32:online} and Yahoo Finance.
We use the metrics from four dimensions: \textbf{Rates \& Policy} includes the US Federal Funds Rate, US 10Y Treasury Yield, JP Policy Rate, and JP 10Y JGB Yield; 
\textbf{Inflation \& Commodities} covers US and JP CPI, along with Gold and Crude Oil prices;
\textbf{Growth \& Economy} tracks US Non-Farm Payrolls, Industrial Production, Housing Starts, Unemployment Rate, and the JP Business Conditions Index; and 
\textbf{Market \& Risk} indicators include the USD/JPY exchange rate, major equity indices (Nikkei 225, S\&P 500), and volatility indices (US VIX, Nikkei VI).
For all indicators, we utilize the latest available closing values and their month-over-month rates of change.

\section{Methodology}
This section details the fine-grained tasks we provide to agents in prompts and the design of experiments to verify their effectiveness.

The core methodological contribution of this study lies in the granularity of instructions (prompts) for AI agents. 
Theoretically, the decomposition of tasks can be arbitrarily granular. 
However, to operationalize automated trading, we define fine-grained tasks as standard analytical tasks that should be performed routinely in practical operations by professional analysts.
While these tasks are fundamental to human practitioners, they have received limited attention in existing financial multi-agent research.

\subsection{Tasks and Prompt Design}
Here, we define the tasks for the seven individual agents.
Among these, the Technical and Quantitative Agents are the targets of our experiments; thus, we describe both fine-grained and coarse-grained tasks for these agents.
We selected these two agents because, among the other agents, only they produce outputs that directly lead to investment execution and their numerical data processing can be clearly defined.
We provide the full prompts in Appendix~\ref{prompts}.

\subsubsection*{1. Technical Agent (Technical Analysis)}
This agent analyzes stock price movement to generate attractive scores ranging from 0 to 100 for each stock.

For \textbf{fine-grained tasks}, we pass pre-calculated technical indicators commonly used in market analysis to the Technical Agent.
To eliminate the bias of nominal price levels, all indicators are normalized or ratio-adjusted. 
We refer existing research on the selection of indicators~\cite[e.g.,][]{xia2023stock,montazeri2025finding,liu2025qtmrl,xiao2024tradingagents,li2022stock,papageorgiou2024enhancing,zhang2024multimodal,xiong2025quantagent}.

\noindent \textbf{Momentum: Rate of Change (RoC):} 
We compute the RoC across multiple lookback horizons (5, 10, 20, and 30 days; 1, 3, 6, and 12 months) to capture trend strength and temporal persistence.

\noindent \textbf{Volatility: Bollinger Band:} Instead of price bands, we identify statistically extreme levels using the Z-score formula $Z = (P - \mu_{20}) / \sigma_{20}$, which normalizes the price deviation from the 20-day moving average by the standard deviation.

\noindent \textbf{Oscillator: Moving Average Convergence Divergence (MACD):} This metric measures trend momentum as the difference between short-term (12-day) and long-term (26-day) exponential moving averages (EMA) of price, with a 9-day EMA of the MACD line as the ``signal line,'' and their difference as the ``histogram.'' 
We show the detailed formulations in Appendix~\ref{formulas}.

\noindent \textbf{Oscillator: Relative Strength Index (RSI):} This metric compares the magnitude of recent gains and losses over a 14-day lookback period to identify potentially overbought or oversold conditions. 
We show the detailed formulations in Appendix~\ref{formulas}.

\noindent \textbf{Oscillator: (KDJ):} We calculate the stochastic oscillator $\%K$ (price position within the 9-day high–low range), $\%D$ (3-day moving average of $\%K$), and the divergence $J = 3D - 2K$ to capture potential trend reversals.
We show the detailed formulations in Appendix~\ref{formulas}.

For \textbf{coarse-grained tasks} for the Technical Agent, instead of adding pre-calculated metrics, we add the raw data directly to the prompts.
Specifically, we feed the data used to calculate the metrics employed in fine-grained tasks (i.e., daily prices over one year) directly into the agent.

\subsubsection*{2. Quantitative Agent (Quantitative Fundamentals)}
This agent quantitatively evaluates a company's financial health and growth potential based on the numbers on financial statements, then generates attractive scores ranging from 0 to 100 for each stock.

For all metrics, we provide the agent with the numerical values and the RoC compared to the previous year, regardless of the granularity of tasks.
For flow variables (e.g., ROE and Sales), we calculate valuation metrics using Trailing Twelve-Month (TTM) to mitigate seasonality while incorporating the most recent performance~\cite{penman2010financial}. 
For stock variables (e.g., Equity Ratio, Total Assets), we utilize values from the latest quarterly balance sheet to ensure timeliness~\cite{livnat2006comparing}.
When a metric cannot be computed due to insufficient data, we pass NaN values to the LLM and rely on the model to handle missing data appropriately.

For \textbf{fine-grained tasks}, as in the Technical Agent, we follow the standard practices in financial analysis~\cite[e.g.,][]{piotroski2000value,lee2025value,lin2019quantitative} to calculate the traditional investment metrics with the five dimensions:
\textbf{Profitability:} ROE, ROA, Operating Profit Margin, FCF Margin;
\textbf{Safety:} Equity Ratio, Current Ratio, D/E Ratio;
\textbf{Valuation:} P/E Ratio, EV/EBITDA Multiple, Dividend Yield;
\textbf{Efficiency:} Total Asset Turnover, Inventory Turnover Period; and 
\textbf{Growth:} Revenue Growth Rate (CAGR), EPS Growth Rate.

For \textbf{coarse-grained tasks} for the Quantitative Agent, instead of providing aggregated financial metrics, we provide the agent with raw data points that we can get from financial statements, such as:
\textbf{Income Statement:} Sales, Operating Profit, Net Income, Cost of Sales, Depreciation;
\textbf{Balance Sheet:} Total Assets, Equity, Cash, Receivables, Financial Assets, Inventory, Current Liabilities, Interest Bearing Debt;
\textbf{Cash Flow:} Operating, Investing; and
\textbf{Market Data:} Monthly Close, EPS, Dividends, Issued Shares.
Additionally, historical EPS data (1-year and 3-year lookbacks) are included to assess long-term earnings stability.

\subsubsection*{3. Qualitative Agent (Qualitative Fundamentals)}
This agent analyzes unstructured text data from securities reports to evaluate sustainability and competitive advantages that are not captured by numerical metrics. 
It extracts information from specific sections and analyzes it based on predetermined evaluation criteria:
\textbf{Business Overview} evaluates business model robustness based on ``History,'' ``Business Description,'' and ``Affiliated Companies;''
\textbf{Risk Analysis} extracts potential downside risks from ``Business Risks'' and ``Issues to Address;''
\textbf{Management Policy} evaluates strategic intent and execution capability based on ``Management's Discussion and Analysis (MD\&A);''
\textbf{Governance} assesses management transparency via board composition (proportion of outside directors) and ``Corporate Governance Status.''
The agent outputs business, risk, and management scores (5-point scale) and reasoning texts as input to the Sector Agent.

\subsubsection*{4. News Agent (News Sentiment and Events)}
This agent aggregates recent news headlines and summaries from major economic media outlets. 
It detects material events such as earnings revisions, scandals, new product announcements, and M\&A activities. 
For each company, the system inputs news data from the current month; if no news is available, we use NaN as input. 
The agent searches for headlines containing company names (and their abbreviations) and extracts the headline and content when found. 
The agent outputs risk and return outlook scores (5-point scale) and reasoning texts as input to the Sector Agent.

\subsubsection*{5. Sector Agent (Sector-Level Adjustment)}
This agent synthesizes the outputs from the four analyst agents and compares the quantitative figures of each stock with the sector averages.
The agent provides the re-evaluated attractive score (0-100 scale) and an investment thesis to the PM agent.

\subsubsection*{6. Macro Agent (Macro-Environmental Assessment)}
This agent analyzes the economic environment with five dimensions---Market Direction, Risk Sentiment, Economic Growth, Interest Rates, and Inflation---based on the absolute levels and month-to-month changes of JP/US economic indicators, then provides the scores (0-100 scale) for each dimension and reasoning texts to the PM agent.  \looseness=-1

\subsubsection*{7. PM Agent (Final Portfolio Construction)}
This agent integrates the bottom-up view (from the Sector Agent) with the top-down view (from the Macro Agent), then generates a final attractive score (0-100 scale) for the long-short portfolio construction.

\section{Experimental Results}
In this section, we present the empirical backtesting results of the proposed multi-agent trading system using the Japanese TOPIX 100 universe from September 2023 to November 2025. 

\subsection{Fine-Grained vs Coarse-Grained Tasks}
To evaluate the effectiveness of fine-grained task decomposition, which is our main objective, we compare the performance of the proposed method (agents with fine-grained tasks) against the baseline method (with coarse-grained tasks) across varying portfolio sizes ($N \in \{10, 20, 30, 40, 50\}$).
Note that $N=10$, for instance, indicates that we long five stocks and short five stocks.
We conduct the experiments in two settings: using all agents and agents without one agent (leave-one-out).

\subsubsection{Comparison using All Agents}

\begin{figure}[!ht]
    \centering
    \includegraphics[width=0.80\linewidth]{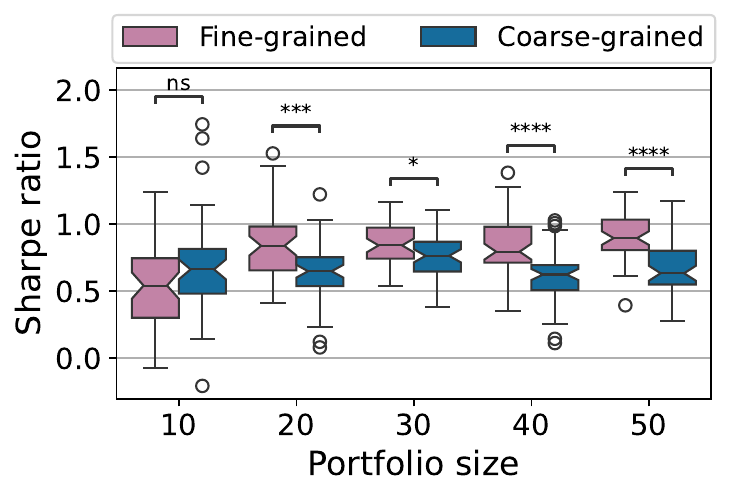}
    \caption{
        Sharpe ratios for fine-grained (pink) and coarse-grained (blue) settings across portfolio sizes. 
        Box plot notches represent the 95\% confidence interval of the median. 
    }
    \label{fig:fine_vs_coarse}
\end{figure}

Figure~\ref{fig:fine_vs_coarse} shows the comparison of Sharpe ratios between fine-grained (pink) and coarse-grained (blue) task settings across five different portfolio sizes, with each configuration evaluated over 50 independent trials.
Stars (*) indicate Mann-Whitney U test significance: $p<$0.0001, 0.001, 0.05 shown as ****, ***,  *.  
`ns' indicates not significant.
Hereafter, we use the same notation throughout the paper. 
As indicated by the stars, the agents with the fine-grained tasks significantly outperform their coarse-grained counterparts in 4 out of the 5 tested horizons (20, 30, 40, and 50 stocks). 
The only exception is the initial case (10 stocks), denoted as "ns" (not significant), which may be attributed to the relatively small number of stocks that renders the backtesting results noisy and unstable. 
Overall, despite the noise in the smallest setting, the aggregate trend demonstrates that providing detailed (fine-grained) information contributes to superior risk-adjusted returns.

\subsubsection{Comparison with Leave-One-Out Settings}
We further compare fine-grained and coarse-grained agent settings under leave-one-out settings. 
Specifically, we systematically remove one of the bottom-level specialist agents---namely, Technical, Quantitative, Qualitative, News, and Macro---to assess the robustness of the proposed method's superiority.
Table~\ref{tab:fine_vs_coarse_ablation} reports the \textit{differences} in median Sharpe ratio between fine-grained and coarse-grained settings, as \\$\Delta SR = \text{Median}(SR_{\text{fine}}) - \text{Median}(SR_{\text{coarse}})$. 
The top row ("All agents") corresponds to the all-agent configuration, consistent with the results shown in Figure~\ref{fig:fine_vs_coarse}.
\begin{table}[!ht]
    \centering
    \caption{Sharpe Ratio Differences Between Fine-grained and Coarse-grained Settings}
    \begin{adjustbox}{max width=\linewidth}
    \begin{tabular}{llllll}
    \toprule
    {Portfolio size} &     \multicolumn{1}{c}{10} & \multicolumn{1}{c}{20} & \multicolumn{1}{c}{30} & \multicolumn{1}{c}{40} & \multicolumn{1}{c}{50} \\
    \midrule
    All agents           &     -0.12 &  +0.19**** &     +0.08* &  +0.17**** &  +0.26**** \\
    w/o Technical &  +0.54*** &      -0.07 &   -0.34*** &  -0.66**** &  -0.79**** \\
    w/o Quant.    &      +0.1 &      +0.04 &      +0.16 &      +0.2* &      +0.12 \\
    w/o Qual.     &  +0.49*** &     +0.24* &   +0.41*** &  +0.55**** &   +0.33*** \\
    w/o News      &    +0.35* &   +1.0**** &  +1.04**** &  +1.04**** &  +1.08**** \\
    w/o Macro     &     +0.11 &       +0.1 &      +0.23 &     +0.31* &      +0.01 \\
    \bottomrule
\end{tabular}
\end{adjustbox}
    \label{tab:fine_vs_coarse_ablation}
\end{table}
Across leave-one-out settings, we find that the differences are predominantly positive in most configurations. 
This indicates that the fine-grained architecture generally achieves higher Sharpe ratios than the coarse-grained baseline, even when specific analytical perspectives are removed. 
A notable exception is the ``w/o Technical'' setting, where the performance reverses for larger portfolio sizes, suggesting that the Technical Agent plays a central role in driving the performance advantage of fine-grained task decomposition.
Overall, the results demonstrate that the fine-grained task design robustly outperforms the equivalent coarse-grained design in backtesting.

\subsection{Ablation Studies}
We conduct ablation studies to quantify the contribution of each specialized agent to overall performance.
Table~\ref{tab:ablation_comparison} presents the changes in the Sharpe ratios compared to the ``All agents'' baseline. 
The values represent the difference calculated as $\text{SR}_{\text{ablation}} - \text{SR}_{\text{baseline}}$. 
Consequently, a positive value indicates that removing the agent improved performance (implying the agent was detrimental or noisy), while a negative value indicates that performance degraded (implying the agent was beneficial).
\begin{table}[!ht]
    \centering
    \caption{Trading performance under Ablation Settings}
    \label{tab:ablation_comparison}
\begin{subtable}{\linewidth}
    \centering
    \caption{Fine-grained settings} 
    \begin{adjustbox}{max width=\linewidth}
        \begin{tabular}{llllll}
            \toprule
            {} &               \multicolumn{1}{c}{10} & \multicolumn{1}{c}{20} & \multicolumn{1}{c}{30} & \multicolumn{1}{c}{40} & \multicolumn{1}{c}{50} \\
            \midrule
            All agents (Baseline) &       \multicolumn{1}{c}{0.54} &      \multicolumn{1}{c}{0.84} &       \multicolumn{1}{c}{0.84} &        \multicolumn{1}{c}{0.79} &        \multicolumn{1}{c}{0.9}\\
            w/o Technical         &      +0.13 &  -0.42**** &   -0.4**** &   -0.56**** &  -0.66**** \\
            w/o Quant.            &  +0.45**** &  +0.23*** &  +0.41**** &  +0.48**** &   +0.2**** \\
            w/o Qual.             &       +0.16 &  +0.27*** &   +0.47**** &   +0.5**** &  +0.33**** \\
            w/o News              &      +0.21 &   +0.23** &   +0.29*** &      +0.12 &  +0.25**** \\
            w/o Macro             &     +0.28* &    +0.15* &     +0.27* &  +0.36**** &   +0.16*** \\
            \bottomrule
        \end{tabular}
    \end{adjustbox}
    \label{tab:ablation_results_fine}
\end{subtable}

    \bigskip  

    \begin{subtable}{\linewidth}
        \centering
        \caption{Coarse-grained settings} 
        \begin{adjustbox}{max width=\linewidth}
            \begin{tabular}{llllll}
                \toprule
                {} &              \multicolumn{1}{c}{10} & \multicolumn{1}{c}{20} & \multicolumn{1}{c}{30} & \multicolumn{1}{c}{40} & \multicolumn{1}{c}{50} \\
                \midrule
                All agents (Baseline) &       \multicolumn{1}{c}{0.66} &       \multicolumn{1}{c}{0.65} &       \multicolumn{1}{c}{0.76} &        \multicolumn{1}{c}{0.62} &        \multicolumn{1}{c}{0.63} \\
                w/o Technical         &  -0.52**** &     -0.16* &      +0.02 &   +0.28*** &   +0.4**** \\
                w/o Quant.            &   +0.24*** &  +0.38**** &  +0.33**** &  +0.46**** &  +0.34**** \\
                w/o Qual.             &  -0.45**** &   +0.22*** &      +0.14 &      +0.13 &  +0.26**** \\
                w/o News              &    -0.26** &  -0.59**** &  -0.68**** &   -0.75**** &  -0.57**** \\
                w/o Macro             &      +0.05 &  +0.23**** &      +0.11 &     +0.22* &   +0.41*** \\
                \bottomrule
            \end{tabular}
        \end{adjustbox}
        \label{tab:ablation_results_coarse}
    \end{subtable}
\end{table}
Overall, most ablation settings show positive differences relative to the full-agent configuration, indicating that most individual agents may introduce noise or redundant signals.
This highlights the importance of carefully designing agent roles and interactions, rather than simply increasing the number of specialized components.
\looseness=-1

In the fine-grained setting (Table~\ref{tab:ablation_comparison}a), however, the ``w/o Technical'' condition shows predominantly negative differences, especially for larger portfolio sizes. 
This implies that the Technical Agent provides particularly strong predictive signals. 
In contrast, removing other agents (Macro, Quantitative, Qualitative) often results in positive differences, suggesting that while they may appear to contribute useful information, they may also introduce noise or redundant signals when combined under fine-grained coordination.

In the coarse-grained setting (Table~\ref{tab:ablation_comparison}b), a similar but weaker pattern is observed for the Technical Agent, further supporting the importance of technical signals in the overall system. 
Notably, the ``w/o News'' row exhibits strongly negative differences across most portfolio sizes, with relatively large magnitudes. 
This behavior is not clearly observed in the fine-grained setting. 
One possible interpretation is that, in the absence of fine-grained task decomposition, news information may be relatively better utilized, potentially compensating for weaker propagation of technical signals.

Overall, the results indicate that performance depends not only on agent diversity but also on how information is structured and propagated across the system. 
Fine-grained task decomposition appears to facilitate more effective signal transmission—particularly for technical signals—while reducing redundancy and noise introduced by loosely coordinated agents. 
This highlights the importance of task design and information routing in hierarchical LLM agent architectures.

\subsection{Text Analysis for Interpretability}
Understanding the rationale behind LLM outputs is critical for practical deployment, especially in financial trading. 
Given the significant performance disparities observed between different prompting strategies in our experiments, we analyze the textual outputs generated during backtesting.

First, we compare the textual outputs generated under fine-grained and coarse-grained settings using the log-odds ratio with a Dirichlet prior~\cite{monroe2008fightin}, a statistical measure to compare how strongly a word is associated with one group versus another.
Second, we analyze information propagation across agents to quantify how much information from lower-level agents is reflected in higher-level agents’ outputs. 
Concretely, inspired by the previous research on information propagation~\cite[e.g.,][]{cann2025using,zhou2025guardian}, we first convert each agent’s output text into vector representations using an LLM embedding model (text-embedding-3-small~\cite{textembe55:online}). 
We then compute cosine similarities between agent output vectors to measure the degree of semantic alignment, which serves as a proxy for information adoption across the agent hierarchy.

\subsubsection{Representative words of Fine-grained vs Coarse-grained Settings}

We obtained the highest log-odds ratios for fine-grained and coarse-grained settings across four agent types: Technical, Quantitative, Sector, and PM Agents.
We present the complete list of words (shown in Table~\ref{tab:log_odds}), as well as the preprocessing of texts, in Appendix~\ref{logodds} for brevity.

First, we observe that both the fine-grained and coarse-grained configuration emphasizes vocabulary that is closely aligned with their respective prompt instruction, which is consistent with our expectations.
The fine-grained setting tends to produce more nuanced analytical terms such as ``momentum,'' ``volatility,'' and ``condition'' for Technical, and ``margins,'' ``growth-rate,'' and ``profitability'' for Quant. 
In contrast, the coarse-grained setting favors more surface-level market descriptors, such as ``price,'' ``trend,'' ``rise,'' and ``increase'' for Technical, and ``EPS,'' ``earnings,'' and ``net income''–related expressions for Quant. 
This confirms that prompt granularity directly influences the level of abstraction in generated reasoning: without explicit procedural guidance, the LLM reverts to broad, superficial descriptions of market movements and financial statements.
\looseness=-1

Second, observing the hierarchical relationships, we find evidence of vocabulary propagation from lower-level specialists to higher-level decision makers. 
Higher-level agents, namely the Sector and PM Agents, tend to reuse or inherit vocabulary originating from lower-level agents. 
For instance, in the fine-grained setting, the PM and Sector Agents' distinctive words include ``Momentum'' (characteristic of the Technical Agent) and ``Soundness'' (characteristic of the Quant Agent). 
Similarly, in the coarse-grained setting, the PM and Sector Agents adopt ``Trend'' and ``EPS,'' mirroring the vocabulary of their respective subordinates. 
This indicates that the hierarchical architecture enables upward propagation of semantic signals, suggesting that higher-level decision-making is at least partially grounded in lower-level analytical outputs.

\subsubsection{Information Propagation Analysis}
Table~\ref{tab:similarity_fine_vs_coarse} shows the cosine similarity between the Sector Agent outputs and those of lower-level agents, which presents the median values of the aggregated results over 50 independent backtesting trials.
The table reports similarities under both fine-grained and coarse-grained settings, along with Diff.\ $=\ \text{Similarity}_{\text{Fine-grained}} - \text{Similarity}_{\text{Coarse-grained}}$.
Regarding the absolute magnitude of similarity, the Quantitative and Qualitative Agents exhibit relatively high scores ($\approx 0.48 - 0.52$) compared to the Technical Agent ($\approx 0.40 - 0.42$).
This suggests that, by default, the Sector Agents' decision-making logic aligns more closely with fundamental analysis (financials and business models) rather than technical price analysis.
One might think that the length of the output of each agent affects the similarity, but we set the output length within 100 Japanese characters for each agent, and we observed no significant difference in the output lengths across agents.
\begin{table}[!ht]
    \centering
    \caption{Semantic Similarity with Sector Agent }
    \begin{adjustbox}{max width=0.8\linewidth}
\begin{tabular}{llll}
\toprule
Agents & Fine-grained & Coarse-grained &      Diff. \\
\midrule
Technical    &        0.419 &          0.397 &  \textbf{0.022} \\
Quantitative &        0.476 &          0.477 & -0.001 \\
Qualitative  &        0.514 &          0.514 & -0.001 \\
News         &        0.378 &          0.372 &  0.006 \\
\bottomrule
\end{tabular}
\end{adjustbox}
    \label{tab:similarity_fine_vs_coarse}
\end{table}
\noindent However, most importantly, regarding the comparison between the fine-grained and coarse-grained settings, we find that only the Technical Agent demonstrates a significant improvement in the fine-grained setting (0.022 in Diff.).
Specifically, in the fine-grained setting, the higher similarity score suggests that technical insights are effectively transmitted and integrated into the Sector Agent's reasoning process.
This aligns with the backtesting results; the Technical Agents perform well, and their impact is critically high, especially in a fine-grained setting.

Note that, across trials, the similarity values exhibit very small variance, with below 0.002 standard deviation for all cases. 
Because of this small deviation, the difference between the two settings (Diff.) is statistically significant in all cases.

Taken together, while the system naturally leans towards fundamental data, the proposed fine-grained architecture effectively amplifies the signal transmission of technical analysis, indicating that technical factors are explicitly integrated into the higher-level decision-making process in the fine-grained setting.

\subsection{Portfolio Optimization}
In regulated financial environments, deploying a fully autonomous trading system directly to live capital is typically infeasible without extensive staged validation.
Consequently, to demonstrate the real-world applicability of our system under these constraints, we perform standard portfolio optimization against a market index as a realistic pre-deployment validation setting.
We conduct a systematic backtest to evaluate the allocation between the TOPIX 100 equity index and a composite portfolio of six LLM-based agent strategies.
Six LLM-based agent strategies include the strategies using all agents and five leave-one-out strategies, constructed using an equal risk contribution weighting scheme. 
Here, we exploit the heterogeneity in outputs across the six strategies---instructed with different combinations of information sources. \looseness=-1

The covariance structure among the six agent strategies is derived from the stock-level covariance matrix of TOPIX 100 constituents. 
Letting $V \in \mathbb{R}^{n \times n}$ denote the stock covariance matrix and $P \in \mathbb{R}^{M \times n}$ the portfolio weight matrix whose rows correspond to each agent's stock-level holdings, the agent-level covariance is computed as $\Sigma = P V P^\top$. 
To combine the six strategies into a single composite, we solve for the weight vector $w$ that equalizes each agent's risk contribution to total portfolio variance: each asset's contribution $w_i (\Sigma w)_i / \sqrt{w^\top \Sigma w}$ is driven toward $1/N$ via constrained optimization, subject to $\sum_i w_i = 1$.

\begin{figure}[!ht]
    \centering
    \includegraphics[width=0.8\linewidth]{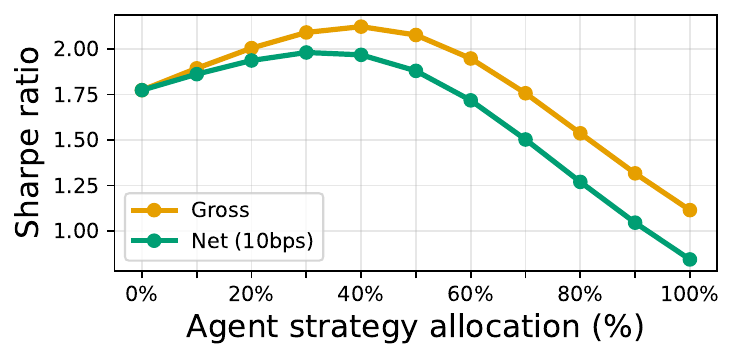}
    \caption{
        Sharpe ratio as a function of allocation between TOPIX 100 and the aggregated agent strategy in the test period. 
        Gross performance (orange) and net performance after 10 bps one-way transaction cost (green) are shown.
    }
    \label{fig:portfolio_optimization}
\end{figure}

We then vary the allocation ratio between TOPIX 100 and the agent composite from $0\%$ to $100\%$ in $10\%$ increments, evaluating out-of-sample performance including annualized return, volatility, and Sharpe ratio, net of realistic transaction costs (10bps one-way).
A key empirical finding is that the correlation between the TOPIX 100 index returns and the agent composite returns is low ($\approx 0.4$), creating substantial diversification benefits. 
As Figure~\ref{fig:portfolio_optimization} shows, blended portfolios consistently achieve higher Sharpe ratios than either the TOPIX 100 index or the agent composite alone (The return and volatility are shown in Table~\ref{tab:oos_performance} in Appendix~\ref{section_performance}). 
While the ex-ante optimal allocation ratio is unknown in practice, even a na\"{i}ve $50\%/50\%$ split between the index and the agent composite yields a Sharpe ratio superior to both standalone components---demonstrating that practitioners can capture meaningful risk-adjusted performance improvements without requiring precise allocation optimization.

\section{Discussion and Conclusion}

In this study, we constructed a hierarchical multi-agent trading framework and investigated how task granularity affects system behavior by comparing fine-grained and coarse-grained task settings. 
Our experimental results demonstrate that the fine-grained setting yields statistically superior overall performance in terms of risk-adjusted returns.
Furthermore, the ablation study revealed that the \textit{Technical} agent is a primary performance driver.
Crucially, text-based analyses consistently confirmed that the fine-grained instructions enabled the effective propagation of technical insights to higher-level decision-makers.
These results provide evidence from performance, ablation, and textual behavior analyses that fine-grained task structuring improves both the effectiveness and the information flow within hierarchical LLM agent systems.
Finally, we evaluated real-world performance through portfolio optimization using market indices as benchmarks.
Our findings suggest that the performance of LLM-based trading agents is driven not merely by the model's reasoning capability, but significantly by the quality of feature engineering embedded within the prompt design.

\subsection{Implications}
From an agent design perspective, our findings suggest a shift in how to construct multi-agent systems for financial analysis. 
Prior work has often implicitly assumed a one-to-one mapping between data modality and agent specialization, with the role-based ambiguous instructions. 
However, our results indicate that agents may be more effectively designed around task decomposition rather than data source boundaries. 
In practical settings, this opens the possibility that users can embed their own domain-specific expertise directly into task-specialized agents, enabling customizable and organization-specific agent frameworks. 

Another important implication concerns interpretability. 
Our study demonstrates that meaningful insights can be extracted from analyzing agent text outputs, providing a practical pathway for understanding LLM-driven decision processes.
Interpretability is especially critical in enterprise settings, particularly for large-scale asset management. 
Prior work has suggested that the adoption of LLMs introduces new operational workflows centered around validating generated outputs~\cite{khojah2025llm}. 
In this context, building trading agents with strong interpretability characteristics is not merely desirable but may become operationally necessary.  \looseness=-1

Simultaneously, there is an ongoing debate over whether natural language should be adopted as the primary communication interface in multi-agent LLM systems. 
Many existing frameworks~\cite{li2023camel,zhuge2023mindstorms,zhang2024proagent,park2023generative} rely on natural language communication, while alternative approaches propose machine-oriented languages that are mutually intelligible among AI agents to improve efficiency and accuracy~\cite{zheng2025thought,xiao2025transmission}. 
Nevertheless, from a practical perspective~\cite{jadhav2025large}, natural language interfaces appear advantageous, as they enable interpretability and downstream analyses such as those conducted in our study. \looseness=-1

\subsection{Limitations and Future Work}
Despite promising results, several limitations remain.
First, it is not yet fully clear whether the performance gains are fundamentally attributable to fine-grained task decomposition itself. 
One alternative explanation is that certain vocabulary patterns may be more easily adopted by the preference of LLMs to influence downstream agents.  
Investigating linguistic bias in LLM-based multi-agent systems is, therefore, an important direction for future research~\cite[c.f.,][]{lee2025your,hwang2026wording}. \looseness=-1

Second, due to the knowledge cutoff of the LLM model, backtesting was limited to approximately two years of historical data. 
Financial markets exhibit strong regime shifts over longer horizons, and therefore, longer-term validation is necessary to confirm robustness. 
One possible future direction is the use of time-aware or temporally constrained LLM variants, such as approaches similar to Time Machine GPT~\cite{drinkall2024time}, which could enable historically consistent simulations across longer market periods.  \looseness=-1

Third, while we established rigorous experimental settings, we acknowledge that there remains room for exploration under different conditions, such as employing other LLM models or targeting different markets (e.g., the US market). 
However, we believe the scope of this analysis is sufficient for the following reasons. 
Primarily, from an industrial perspective, our primary focus is on the Japanese market due to specific deployment requirements. 
Also, current LLMs have demonstrated sufficient pre-trained knowledge regarding the Japanese market~\cite{okada2025words}.
Furthermore, given the recent competitive progress across providers~\cite{LLMBench11:online}, we infer that the fundamental validity of our proposed method remains consistent across high-performing models.


\bibliographystyle{ACM-Reference-Format}

\appendix

\section{Formulas for Technical Indicators}\label{formulas}
\subsection{MACD}
We calculate the MACD line ($M_t = \text{EMA}_{12} - \text{EMA}_{26}$), the signal line ($S_t = \text{EMA}_{9}(M_t)$), and the histogram ($H_t = M_t - S_t$), where EMA indicates the exponential moving average with the smoothing factor $\alpha = \frac{2}{t+1}$. 
These values are normalized by the closing price $P_t$ (i.e., $M_t / P_t$) to enable cross-sectional comparison.

\subsection{RSI}
The RSI is defined as
\[
RSI_t = 100 - \frac{100}{1 + RS_t},
\]
where
\[
RS_t = \frac{\text{AvgGain}_t}{\text{AvgLoss}_t}.
\]
The average gain and loss are computed using exponentially smoothed moving averages over a 14-day lookback period.

\subsection{Stochastic Oscillator Formulation}

Let $P_t$ denote the closing price at time $t$.  
We define the highest and lowest prices over the past $n$ days as
\begin{align}
H_t^{(n)} &= \max_{i=0,\dots,n-1} P_{t-i}, \\
L_t^{(n)} &= \min_{i=0,\dots,n-1} P_{t-i}.
\end{align}

The stochastic oscillator $\%K$ is defined as
\begin{align}
\%K_t = 100 \times \frac{P_t - L_t^{(9)}}{H_t^{(9)} - L_t^{(9)}}.
\end{align}

The signal line $\%D$ is computed as the 3-day simple moving average (SMA) of $\%K$:
\begin{align}
\%D_t = \frac{1}{3} \sum_{i=0}^{2} \%K_{t-i}.
\end{align}

Finally, the divergence term $J$ is defined as
\begin{align}
J_t = 3\%D_t - 2\%K_t.
\end{align}

\section{Prompts}\label{prompts}

\subsection{Technical Agent}

\subsubsection{System Prompt}

\begin{compactprompt}
\textbf{System Prompt (Role Definition):}

\textbf{Role:} You are a technical analyst on the trading team. Your task is to forecast stock prices one month ahead based strictly on technical indicators to assist portfolio managers.

\vspace{0.5em}
\textbf{Policy \& Constraints:}
\begin{itemize}[leftmargin=1.5em, nosep, label={-}]
    \item \textbf{Input scope:} Use \textit{only} the provided technical indicators; disregard news or fundamentals.
    \item \textbf{Scoring:} Provide a score between 0 and 100 based on a balanced assessment of momentum, oscillators, and volatility.
    \item \textbf{Scale interpretation:} 
    \begin{itemize}[nosep, label={$\cdot$}]
        \item \textbf{100}: Strong long recommendation
        \item \textbf{50}: Neutral (no clear advantage)
        \item \textbf{0}: Strong short recommendation
    \end{itemize}
    \item \textbf{Output requirements:} strictly JSON format, including a brief one-sentence comment.
\end{itemize}
\end{compactprompt}


\subsubsection{User Prompt (Fine-Grained)}
\begin{compactprompt}
\textbf{Instruction:}
The following are technical indicators for a particular stock at the end of a given month. Based on these, please rate the attractiveness of long or short positions in this stock on a scale of 0 to 100 points.

\vspace{0.5em}
\textbf{Technical indicators used (summary of definitions/reference):}
\begin{itemize}[leftmargin=1.5em, nosep, label={-}]
    \item \textbf{Momentum: RoC (Rate of Change)}
    \begin{itemize}[nosep, label={$\cdot$}]
        \item Percentage change in price over the past $n$ days ($n = 5, 10, 20$, compared to previous month) and $m$ months ($n = 1, 3, 6, 12$).
    \end{itemize}
    \item \textbf{Volatility: Bollinger Band Deviation}
    \begin{itemize}[nosep, label={$\cdot$}]
        \item Actual value used: (Close - 20-day MA) / 20-day Close standard deviation
    \end{itemize}
    \item \textbf{Oscillators:}
    \begin{itemize}[nosep, label={$\cdot$}]
        \item \textbf{MACD:} (12-day EMA - 26-day EMA) / Close price. Normalized by dividing by Close.
        \item \textbf{RSI:} Ranges from 0 to 100. (100 $\times$ Upward Avg) / (Upward Avg + Downward Avg).
        \item \textbf{Stochastic Oscillator:} Uses K, D, J indicators.
    \end{itemize}
\end{itemize}

\vspace{0.5em}
\textbf{Evaluation Rules:}
\begin{itemize}[leftmargin=1.5em, nosep, label={-}]
    \item Comprehensively assess the risk-reward ratio for the next month based on the combination of indicators.
    \item Check consistency among momentum, oscillators, and volatility.
    \item \textbf{100}: Strong Long, \textbf{0}: Strong Short, \textbf{50}: Neutral.
    \item State the reason in Japanese in one sentence ($\approx$ 50 chars).
\end{itemize}

\vspace{0.5em}
\textbf{Output Format (JSON only):}
\begin{verbatim}
{
  "score": <integer 0-100>,
  "reason": "<explanation>"
}
\end{verbatim}

\vspace{0.5em}
\textbf{This Month's Technical Indicators:}
\begin{itemize}[leftmargin=1.5em, nosep, label={}]
    \item RoC 5day: \texttt{<value>}\% \quad RoC 10day: \texttt{<value>}\%
    \item RoC 20day: \texttt{<value>}\% \quad RoC 1Month: \texttt{<value>}\%
    \item RoC 3Month: \texttt{<value>}\% \quad RoC 6Month: \texttt{<value>}\%
    \item RoC 12Month: \texttt{<value>}\%
    \item RSI: \texttt{<value>}
    \item MACD: \texttt{<value>} \quad Signal: \texttt{<value>} \quad Hist: \texttt{<value>}
    \item Stochastic \%K: \texttt{<value>} \quad \%D: \texttt{<value>} \quad \%J: \texttt{<value>}
\end{itemize}
\end{compactprompt}

\subsubsection{User Prompt (Coarse-Grained)}
\begin{compactprompt}
\textbf{Instruction:}
The following list contains raw daily closing prices for a particular stock over the past 252 business days. The values on the left are the most recent. Based on these, rate the attractiveness of long or short positions on a scale of 0 to 100.

\vspace{0.5em}
\textbf{Evaluation Rules:}
\begin{itemize}[leftmargin=1.5em, nosep, label={-}]
    \item Analyze the price trend to assess the risk-reward ratio for the next month.
    \item \textbf{100}: Strong Long, \textbf{50}: Neutral, \textbf{0}: Strong Short.
    \item State the reason in Japanese ($\approx$ 50 chars).
\end{itemize}

\vspace{0.5em}
\textbf{Output Format (JSON only):}
\begin{verbatim}
{
  "score": <integer 0-100>,
  "reason": "<explanation>"
}
\end{verbatim}

\vspace{0.5em}
\textbf{This Month's Stock Prices (List of 252 days):}
\begin{itemize}[leftmargin=1.5em, nosep, label={}]
    \item \texttt{[<price\_t>, <price\_t-1>, ..., <price\_t-251>]}
    \item (e.g., \texttt{[1500.5, 1498.2, ..., 1200.0]})
\end{itemize}
\end{compactprompt}


\subsection{Quant. Agent}

\subsubsection{System Prompt}
\begin{compactprompt}
\textbf{System Prompt (Role Definition):}

\textbf{Role:} You are a Quantitative Fundamental Analyst. Your task is to evaluate the medium-to-long-term investment attractiveness of a stock based strictly on quantitative financial metrics to assist the Portfolio Manager.

\vspace{0.5em}
\textbf{Guidelines \& Constraints:}
\begin{itemize}[leftmargin=1.5em, nosep, label={-}]
    \item \textbf{Input scope:} Use \textit{only} the provided financial metrics; exclude news, sentiment, or technical patterns.
    \item \textbf{Evaluation Balance:} Assess Profitability (Margins, ROE), Value (PER), Financial Health (Quick Ratio, D/E), Growth, and Cash Flow quality.
    \item \textbf{Scoring Scale:} 
    \begin{itemize}[nosep, label={$\cdot$}]
        \item \textbf{100}: Strong Long (extremely attractive)
        \item \textbf{50}: Neutral (fairly valued)
        \item \textbf{0}: Strong Short (extremely unattractive)
    \end{itemize}
    \item \textbf{Missing Data:} Ignore items marked "NaN" or blank; analyze based on remaining data.
    \item \textbf{Output requirements:} Strictly JSON format; "reason" must be a single short sentence in Japanese.
\end{itemize}
\end{compactprompt}


\subsubsection{User Prompt (Fine-Grained)}
\begin{compactprompt}
\textbf{Instruction:}
The following are fundamental metrics and their changes from a month ago. Evaluate the attractiveness for a Long/Short position on a scale of 0 to 100 based on these.

\vspace{0.5em}
\textbf{Rules \& Definitions:}
\begin{itemize}[leftmargin=1.5em, nosep, label={-}]
    \item \textbf{Metrics:} Profitability, Value (PER), Cash Flow, Financial Health, Growth (Sales/EPS).
    \item \textbf{Trend Analysis:} Consider both "Absolute Value" and "Diff" (change from prev. month).
    \item \textbf{Freshness:} If "Information Update Month" is "Yes", latest results are reflected.
    \item \textbf{Output:} Score (0-100) and reason (in Japanese, $\approx$ 50 chars).
\end{itemize}

\vspace{0.5em}
\textbf{Output Format (JSON only):}
\begin{verbatim}
{
  "score": <integer 0-100>,
  "reason": "<explanation>"
}
\end{verbatim}

\vspace{0.5em}
\textbf{Stock Data (TTM) [Format: Value (diff: Value)]:}
\begin{itemize}[leftmargin=1.5em, nosep, label={}]
    \item \textbf{Info Update Month:} \texttt{<Yes/No>}
    \item \textbf{[Profitability]} Net Margin: \texttt{<val>} (diff: \texttt{<val>}) \quad ROA: \texttt{<val>} (diff: \texttt{<val>}) \\
    ROE: \texttt{<val>} (diff: \texttt{<val>}) \quad Asset Turn: \texttt{<val>} (diff: \texttt{<val>}) \\
    Inv. Turn Days: \texttt{<val>} (diff: \texttt{<val>})
    
    \item \textbf{[Value]} PER: \texttt{<val>} (diff: \texttt{<val>})
    
    \item \textbf{[Cash Flow]} FCF: \texttt{<val>} (diff: \texttt{<val>}) \quad Margin: \texttt{<val>} (diff: \texttt{<val>}) \\
    EBITDA: \texttt{<val>} (diff: \texttt{<val>})
    
    \item \textbf{[Health]} Equity Ratio: \texttt{<val>} (diff: \texttt{<val>}) \quad Quick Ratio: \texttt{<val>} (diff: \texttt{<val>}) \\
    D/E Ratio: \texttt{<val>} (diff: \texttt{<val>})
    
    \item \textbf{[Growth]} Sales YoY: \texttt{<val>} (diff: \texttt{<val>}) \quad CAGR 3Y: \texttt{<val>} (diff: \texttt{<val>}) \\
    EPS Growth: \texttt{<val>} (diff: \texttt{<val>}) \quad DPS: \texttt{<val>} (diff: \texttt{<val>})
\end{itemize}
\end{compactprompt}


\subsubsection{User Prompt (Coarse-Grained)}
\begin{compactprompt}
\textbf{Instruction:}
Evaluate the attractiveness of this stock for a Long/Short position (0-100) based on the fundamental metrics and their changes (RoC) from a month ago.

\vspace{0.5em}
\textbf{Rules \& Format:}
\begin{itemize}[leftmargin=1.5em, nosep, label={-}]
    \item \textbf{Trend Analysis:} Consider both "Absolute Value" and "RoC" (Rate of Change).
    \item \textbf{Handling Missing Data:} Judge based on available info.
    \item \textbf{Output:} JSON format with a score (0-100) and a Japanese reason ($\approx$ 50 chars).
\end{itemize}

\vspace{0.5em}
\textbf{Stock Data (TTM) [Format: Value (RoC: Value\%)]:}
\begin{itemize}[leftmargin=1.5em, nosep, label={}]
    \item \textbf{Info Update:} \texttt{<Yes/No>}
    
    \item \textbf{[P/L]} Sales: \texttt{<val>} (RoC: \texttt{<val>}) \quad Cost of Sales: \texttt{<val>} (RoC: \texttt{<val>}) \\
    Op Profit: \texttt{<val>} (RoC: \texttt{<val>}) \quad Net Income: \texttt{<val>} (RoC: \texttt{<val>}) \\
    Depreciation: \texttt{<val>} (RoC: \texttt{<val>})
    
    \item \textbf{[EPS]} Current: \texttt{<val>} (RoC: \texttt{<val>}) \quad 1y Ago: \texttt{<val>} \quad 3y Ago: \texttt{<val>}
    
    \item \textbf{[B/S: Assets]} Total Assets: \texttt{<val>} (RoC: \texttt{<val>}) \quad Cash: \texttt{<val>} (RoC: \texttt{<val>}) \\
    Receivables: \texttt{<val>} (RoC: \texttt{<val>}) \quad Inventory: \texttt{<val>} (RoC: \texttt{<val>}) \\
    Financial Assets: \texttt{<val>} (RoC: \texttt{<val>})
    
    \item \textbf{[B/S: Liab/Eq]} Equity: \texttt{<val>} (RoC: \texttt{<val>}) \quad Debt: \texttt{<val>} (RoC: \texttt{<val>}) \\
    Cur. Liabilities: \texttt{<val>} (RoC: \texttt{<val>})
    
    \item \textbf{[Cash Flow]} Op CF: \texttt{<val>} (RoC: \texttt{<val>}) \quad Inv CF: \texttt{<val>} (RoC: \texttt{<val>})
    
    \item \textbf{[Others]} Dividends: \texttt{<val>} (RoC: \texttt{<val>}) \quad Issued Shares: \texttt{<val>} (RoC: \texttt{<val>}) \\
    Monthly Close: \texttt{<val>} (RoC: \texttt{<val>})
\end{itemize}
\end{compactprompt}

\subsection{Qual. Agent}
\subsubsection{System Prompt}
\begin{compactprompt}
\textbf{System Prompt (Role Definition):}

\textbf{Role:} You are a Strategic Analyst reporting to the Portfolio Manager. Your mission is to analyze qualitative corporate disclosures and provide a "Fundamental Risk \& Catalyst Report" for the upcoming 1-month horizon.

\vspace{0.5em}
\textbf{Perspective \& Analysis Logic:}
\begin{itemize}[leftmargin=1.5em, nosep, label={-}]
    \item \textbf{Filter:} Distinguish between "stagnant boilerplate text" and "meaningful strategic shifts."
    \item \textbf{Focus:} Identify qualitative triggers (catalysts or red flags) rather than just long-term value.
    \item \textbf{Target:} Operational momentum, management credibility, and hidden structural risks.
\end{itemize}

\vspace{0.5em}
\textbf{Guidelines:}
\begin{itemize}[leftmargin=1.5em, nosep, label={-}]
    \item \textbf{Inputs:} Excerpts from Securities Reports (Business Overview, Risks, MD\&A, Governance).
    \item \textbf{Outputs:} Three specific scores (1-5) and a strategic summary ("Insight").
    \item \textbf{Format:} Return ONLY a JSON object. The "insight" must be written in Japanese.
\end{itemize}
\end{compactprompt}


\subsubsection{User Prompt}

\begin{compactprompt}
\textbf{Instruction:}
Evaluate qualitative corporate data to advise the PM on stock attractiveness and potential risks for the next 1 month.

\vspace{0.5em}
\textbf{Evaluation Items (Score 1-5):}
\begin{enumerate}[leftmargin=1.5em, nosep, label=\textbf{\arabic*.}]
    \item \textbf{Business Momentum:} Strength of cycle/strategy. \\
    (1: Deteriorating/Vague $\to$ 5: Strong tailwinds/Clear execution)
    \item \textbf{Immediate Risk Severity:} Probability of risks manifesting. \\
    (1: High risk/Urgent $\to$ 5: Low risk/Stable)
    \item \textbf{Management Trust:} Credibility \& oversight structure. \\
    (1: Untrustworthy $\to$ 5: Transparent/Aligned)
\end{enumerate}

\vspace{0.5em}
\textbf{Rules \& Output:}
\begin{itemize}[leftmargin=1.5em, nosep, label={-}]
    \item \textbf{Focus:} Look for "Changes" in tone or new risk factors.
    \item \textbf{Insight:} Professional briefing in Japanese ($\approx$ 150 chars).
    \item \textbf{Format:} JSON with scores and insight.
\end{itemize}

\vspace{0.5em}
\textbf{Input Data (Text Excerpts):}
\begin{itemize}[leftmargin=1.5em, nosep, label={}]
    \item \textbf{Info Update:} \texttt{<Yes/No>}
    \item \textbf{[1. Overview]} \texttt{<Business Description text...>}
    \item \textbf{[2. Risks]} \texttt{<Business Risks text...>}
    \item \textbf{[3. MD\&A]} \texttt{<Financial Analysis text...>}
    \item \textbf{[4. Governance]} \texttt{<Officers/Board text...>}
\end{itemize}
\end{compactprompt}

\subsection{News Agent}

\subsubsection{System Prompt}
\begin{compactprompt}
\textbf{System Prompt (Role Definition):}

\textbf{Role:} You are a Senior News Analyst specializing in the stock market. Your task is to analyze news headlines and summaries from the past month to provide qualitative insights that complement fundamental scores.

\vspace{0.5em}
\textbf{Evaluation Guidelines:}
\begin{itemize}[leftmargin=1.5em, nosep, label={-}]
    \item \textbf{Perspectives:} Evaluate impact on "Return Outlook" (Upside) and "Risk Outlook" (Downside).
    \item \textbf{Scoring Scale (1-5):}
    \begin{itemize}[nosep, label={$\cdot$}]
        \item 1: Minimal/None $\to$ 3: Moderate $\to$ 5: Extreme
    \end{itemize}
    \item \textbf{Analysis Logic:} Distinguish between temporary noise and structural changes (e.g., product launches, regulations, ESG).
    \item \textbf{Output:} JSON format only. Reason must be a concise Japanese summary.
\end{itemize}
\end{compactprompt}


\subsubsection{User Prompt}
\begin{compactprompt}
\textbf{Instruction:}
Evaluate "Return Outlook" and "Risk Outlook" (1-3 months) based on the provided news articles.

\vspace{0.5em}
\textbf{Evaluation Criteria (Score 1-5):}
\begin{itemize}[leftmargin=1.5em, nosep, label={-}]
    \item \textbf{Return Outlook:} Positive momentum (e.g., new products, expansion).
    \item \textbf{Risk Outlook:} Potential downside/uncertainty (e.g., supply chain, lawsuits).
\end{itemize}

\vspace{0.5em}
\textbf{Rules \& Output:}
\begin{itemize}[leftmargin=1.5em, nosep, label={-}]
    \item \textbf{Balance:} Identify risks even if news is generally positive.
    \item \textbf{Empty Case:} If no news, set both scores to 1 and reason "No News".
    \item \textbf{Format:} JSON with scores (1-5) and Japanese reason ($\approx$ 100 chars).
\end{itemize}

\vspace{0.5em}
\textbf{News List for the Month (Input Data):}
\begin{itemize}[leftmargin=1.5em, nosep, label={}]
    \item \texttt{<List of [Date] Headline / Summary...>}
    \item (e.g., \texttt{2024-01-15: Launched new EV model...})
    \item (e.g., \texttt{2024-01-20: CEO announced resignation...})
\end{itemize}
\end{compactprompt}

\subsection{Sector Agent}

\subsubsection{System Prompt}

\begin{compactprompt}
\textbf{System Prompt (Role Definition):}

\textbf{Role:} You are a Sector Specialist on the investment committee. Your task is to synthesize reports from Technical, Quantitative, and Qualitative sub-analysts to provide a definitive 1-month investment recommendation.

\vspace{0.5em}
\textbf{Synthesis Logic \& Perspective:}
\begin{itemize}[leftmargin=1.5em, nosep, label={-}]
    \item \textbf{The "Bridge":} Connect raw multi-angle analysis to PM execution.
    \item \textbf{Metaphor:} Tech/Quant act as the "Engine" (price/value); Qualitative acts as the "Steering" (hazards).
    \item \textbf{Dynamic Weighting:} Adjust weights based on consistency and sector environment (e.g., high volatility $\to$ prioritize risk).
\end{itemize}

\vspace{0.5em}
\textbf{Guidelines:}
\begin{itemize}[leftmargin=1.5em, nosep, label={-}]
    \item \textbf{Sector Context:} Compare metrics against sector averages to identify Leaders vs. Laggards.
    \item \textbf{Output:} Final Conviction Score (0-100) and a comprehensive Investment Thesis.
\end{itemize}
\end{compactprompt}


\subsubsection{User Prompt}
We alter the prompts depending on the granularity of settings.

\begin{compactprompt}
\textbf{Instruction:}
As the Sector Specialist, review the analyst reports and sector data to provide a final recommendation (Conviction Score \& Investment Thesis).

\vspace{0.5em}
\textbf{1. Sub-Analyst Reports (Inputs):}
\begin{itemize}[leftmargin=1.5em, nosep, label={-}]
    \item \textbf{Technical Analyst:} \texttt{<Score \& Comment>}
    \item \textbf{Quant Fundamental Analyst:} \texttt{<Score \& Comment>}
    \item \textbf{Qualitative Strategic Analyst:} \texttt{<Score \& Comment>}
\end{itemize}

\vspace{0.5em}
\textbf{2. Sector \& Comparative Context (Variable Inputs):}
\textit{The inputs below switch between Coarse-grained and Fine-grained settings.}

\begin{description}[leftmargin=0.5em, style=unboxed]
    \item[\textbf{[Setting A: Coarse-grained (Financial Ratios)]}] \hfill \\
    Compare Target vs. Sector Avg using high-level metrics:
    \begin{itemize}[leftmargin=1.5em, nosep, label={$\cdot$}]
        \item \textbf{Profitability:} Net Margin, ROA, ROE, Asset Turnover, Inv. Turn Days.
        \item \textbf{Value:} PER.
        \item \textbf{Cash Flow:} FCF, FCF Margin, EBITDA.
        \item \textbf{Health:} Equity Ratio, Quick Ratio, D/E Ratio.
        \item \textbf{Growth:} Sales YoY/CAGR, EPS YoY/3y-Ago, DPS.
    \end{itemize}

    \vspace{0.3em}
    \item[\textbf{[Setting B: Fine-grained (Raw RoC)]}] \hfill \\
    Compare Target vs. Sector Avg using Rate of Change (RoC) for specific items:
    \begin{itemize}[leftmargin=1.5em, nosep, label={$\cdot$}]
        \item \textbf{P/L Items:} Sales, Op Profit, Net Income, Cost of Sales, Depreciation.
        \item \textbf{B/S Items:} Total Assets, Equity, Cash, Receivables, Inventory, Financial Assets, Interest Bearing Debt, Cur. Liabilities, Issued Shares.
        \item \textbf{CF \& Others:} Op CF, Investing CF, Dividends, Monthly Close.
    \end{itemize}
\end{description}

\vspace{0.5em}
\textbf{3. Tasks for PM Report (Output):}
\begin{enumerate}[leftmargin=1.5em, nosep, label=\textbf{\arabic*.}]
    \item \textbf{Conviction Score (0-100):} Integrate views \& sector strength. (100: Outperform, 0: Underperform).
    \item \textbf{Comprehensive Thesis:} Synthesize alignment/conflict between Tech/Fund/Sector. Highlight catalysts/risks. ($\approx$ 200 words in Japanese).
\end{enumerate}

\vspace{0.5em}
\textbf{Output Format:} JSON with \texttt{"score"} and \texttt{"investment\_thesis"}.
\end{compactprompt}

\subsection{Macro Agent}

\subsubsection{System Prompt}
\begin{compactprompt}
\textbf{System Prompt (Role Definition):}

\textbf{Role:} You are a Macro Analyst on the trading team. Your task is to analyze JP/US macro indicators to identify factors influencing the 1-month return of Japanese equities.

\vspace{0.5em}
\textbf{Evaluation Areas (Label \& Score 0-100):}
\begin{itemize}[leftmargin=1.5em, nosep, label={-}]
    \item \textbf{Market Direction:} Bullish/Bearish (Overall outlook).
    \item \textbf{Risk:} Risk sentiment and potential volatility.
    \item \textbf{Economy:} Economic growth trends.
    \item \textbf{Rates:} Interest rate environment.
    \item \textbf{Inflation:} Price level trends.
\end{itemize}

\vspace{0.5em}
\textbf{Policy \& Constraints:}
\begin{itemize}[leftmargin=1.5em, nosep, label={-}]
    \item \textbf{Input Scope:} Use \textit{only} provided indicators; do not interpret news.
    \item \textbf{Scoring Logic:} Based on "Levels" and "Rate of Change" of indicators.
    \item \textbf{Output Requirement:} Strictly JSON format.
    \item \textbf{Summary:} Concise comment in Japanese ($\approx$ 200 chars).
\end{itemize}
\end{compactprompt}


\subsubsection{User Prompt}

\begin{compactprompt}
\textbf{Instruction:}
Evaluate the current macroeconomic environment to impact the 1-month forward return of Japanese stocks, based on "Levels" and "RoC" (Rate of Change).

\vspace{0.5em}
\textbf{Evaluation Items \& Scoring Rules (0-100):}
\textit{Score 100: Strong Buy (Bullish), 50: Neutral, 0: Strong Sell (Bearish).}
\begin{itemize}[leftmargin=1.5em, nosep, label={-}]
    \item \textbf{Market Trend:} Stock indices momentum. (High: Upward trend).
    \item \textbf{Risk Environment:} VIX \& Safe assets. (High Score: Low VIX/Stable Risk-on).
    \item \textbf{Economic Growth:} Employment, Production. (High: Expansion).
    \item \textbf{Interest Rates:} Levels \& Direction. (High Score: Accommodative/Falling).
    \item \textbf{Inflation:} Prices \& Commodities. (High Score: Stable/Disinflation. Low Score: Stagflation/Deflation).
\end{itemize}

\vspace{0.5em}
\textbf{Output Format (JSON only):}
\begin{verbatim}
{
  "metrics": { "market_trend": {"label": "...", "score": 0-100}, ... },
  "summary": "<Implications for active management (approx 200 chars)>"
}
\end{verbatim}

\vspace{0.5em}
\textbf{Macro Indicators [Format: Value (RoC: Value\%)]:}
\begin{itemize}[leftmargin=1.5em, nosep, label={}]
    \item \textbf{[1. Rates \& Policy]} \\
    US Fed Rate: \texttt{<val>} (RoC: \texttt{<val>}) \quad US 10Y Yield: \texttt{<val>} (RoC: \texttt{<val>}) \\
    JP Policy Rate: \texttt{<val>} (RoC: \texttt{<val>}) \quad JP 10Y Yield: \texttt{<val>} (RoC: \texttt{<val>})

    \item \textbf{[2. Inflation \& Commodities]} \\
    US CPI: \texttt{<val>} (RoC: \texttt{<val>}) \quad JP CPI: \texttt{<val>} (RoC: \texttt{<val>}) \\
    Gold: \texttt{<val>} (RoC: \texttt{<val>}) \quad Crude Oil: \texttt{<val>} (RoC: \texttt{<val>})

    \item \textbf{[3. Growth \& Economy]} \\
    US Payrolls: \texttt{<val>} (RoC: \texttt{<val>}) \quad Ind. Prod: \texttt{<val>} (RoC: \texttt{<val>}) \\
    Housing Starts: \texttt{<val>} (RoC: \texttt{<val>}) \quad Unemp. Rate: \texttt{<val>} (RoC: \texttt{<val>}) \\
    JP Business Index: \texttt{<val>} (RoC: \texttt{<val>})

    \item \textbf{[4. Market \& Risk]} \\
    USD/JPY: \texttt{<val>} (RoC: \texttt{<val>}) \quad Nikkei 225: \texttt{<val>} (RoC: \texttt{<val>}) \\
    S\&P 500: \texttt{<val>} (RoC: \texttt{<val>}) \quad US VIX: \texttt{<val>} (RoC: \texttt{<val>}) \\
    Nikkei VI: \texttt{<val>} (RoC: \texttt{<val>})
\end{itemize}
\end{compactprompt}


\subsection{PM Agent}

\subsubsection{System Prompt}
\begin{compactprompt}
\textbf{System Prompt (Role Definition):}

\textbf{Role:} You are the Chief Portfolio Manager (PM). Your task is to determine the definitive investment score by integrating \textbf{Top-Down Macro analysis} with \textbf{Bottom-Up Sector/Stock analysis}.

\vspace{0.5em}
\textbf{Decision Logic (Integration Strategy):}
\begin{itemize}[leftmargin=1.5em, nosep, label={-}]
    \item \textbf{Goal:} Maximize alpha (1-month horizon) while strictly managing risk.
    \item \textbf{Harmonization:} Balance "Market Context" (Macro) vs. "Stock Specifics".
    \item \textbf{Macro Alignment:} If Macro is "Risk-Off", apply a conservative discount (lower scores) unless the stock has exceptional defensive qualities.
    \item \textbf{Tie-Breaking:} Use Macro context to resolve conflicts between Technicals and Fundamentals (e.g., High Inflation $\to$ favor fundamental pricing power over momentum).
\end{itemize}

\vspace{0.5em}
\textbf{Output Requirements:}
\begin{itemize}[leftmargin=1.5em, nosep, label={-}]
    \item \textbf{Final Score:} 0-100 (Definitive conviction).
    \item \textbf{Rationale:} Decisive professional rationale justifying the integration.
\end{itemize}
\end{compactprompt}

\subsubsection{User Prompt}

\begin{compactprompt}
\textbf{Instruction:}
As the Portfolio Manager, review the Macro and Sector-level inputs to provide your final investment decision for the next 1 month.

\vspace{0.5em}
\textbf{Input Reports (Integration):}
\begin{itemize}[leftmargin=1.5em, nosep, label={}]
    \item \textbf{[1. Macro Environment Report]}
    \item \texttt{<Macro Analyst Output (JSON)>}
    \item \textbf{[2. Sector Specialist Report]}
    \item \texttt{<Sector Specialist Output (JSON)>}
\end{itemize}

\vspace{0.5em}
\textbf{Final Decision Tasks:}
\begin{enumerate}[leftmargin=1.5em, nosep, label=\textbf{\arabic*.}]
    \item \textbf{Final Investment Score (0-100):}
    \begin{itemize}[nosep, label={$\cdot$}]
        \item \textbf{100:} Maximum Overweight (High conviction, perfect alignment).
        \item \textbf{50:} Neutral/Hold (Market-weight, no clear edge).
        \item \textbf{0:} Strong Underweight/Avoid (High risk).
    \end{itemize}
    \item \textbf{Final Rationale (Japanese):}
    \begin{itemize}[nosep, label={$\cdot$}]
        \item Explain macro influence on this specific stock.
        \item Summarize key reasons \& risk-reward balance (30 days).
    \end{itemize}
\end{enumerate}

\vspace{0.5em}
\textbf{Output Format (JSON only):}
\begin{verbatim}
{
  "final_score": <integer 0-100>,
  "reason": "<Decisive rationale in Japanese (150-200 chars)>"
}
\end{verbatim}
\end{compactprompt}


\section{Performance by Portfolio Optimization}\label{section_performance}
Table~\ref{tab:oos_performance} shows the detailed performance by optimizing the portfolios constructed with the agents' outputs and index (TOPIX 100).
In the table, return and volatility are annualized, and the Sharpe Ratio assumes a risk-free rate of 0\%. 
Agent Strategies refers to the Risk Parity portfolio combining six LLM-based agent strategies. 
Transaction cost is applied one-way at 10bps per trade for Agent Strategies.

\begin{table}[!htbp]
\centering
\caption{Performance Comparison}
\label{tab:oos_performance}
\begin{adjustbox}{max width=\linewidth}
\begin{tabular}{l|ccc|ccc}
\toprule
Portfolio 
& \multicolumn{3}{c|}{Gross} 
& \multicolumn{3}{c}{Net (10bps)} \\
\cmidrule(lr){2-4} \cmidrule(lr){5-7}
& Ret. & Vol. & S.R. & Ret. & Vol. & S.R. \\
\midrule
TOPIX 100 
& 19.3\% & 11.5\% & 1.68 
& 19.3\% & 11.5\% & 1.68 \\
Agent Strategies 
& 13.7\% & 11.2\% & 1.22 
& 10.6\% & 11.2\% & 0.95 \\
50-50 Combined
& 16.8\% & 8.0\% & \textbf{2.11} 
& 15.2\% & 8.0\% & \textbf{1.91} \\
\bottomrule
\end{tabular}
\end{adjustbox}
\end{table}
    
\section{Representative Words in Outputs of Agents}\label{logodds}

Regarding the log-odds analysis, we first aggregate all text outputs and perform morphological analysis on them. 
We use the Japanese morphological analyzer MeCab~\cite{kudo2005mecab} and the dictionary NEologd~\cite{sato2017mecabipadicneologdnlp2017}. 
From the texts, we extracted only nouns without basic stopwords, created a corpus by fine-grained and coarse-grained settings, and applied log-odds analysis.
Table~\ref{tab:log_odds} shows the top 10 words for each group. 
The words in the table are originally in Japanese but are shown translated into English.

\begin{table*}[!htbp]
    \caption{Top 10 representative words by Log-Odds Ratio with Dirichlet Prior}
    \label{tab:log_odds}
    \begin{tabular}{p{1.2cm}p{7.6cm}p{7.6cm}}
    \toprule
    Agent& \multicolumn{1}{c}{Fine-grained} & \multicolumn{1}{c}{Coarse-grained } \\
    \midrule
    Technical (Level 1) & Momentum, Neutral, Short-term, Favorable, Implication, Suggestion, Long-term, Condition, Decline, Volatility                                & Price, Trend, Rise, Upward-trend, Fall, Stock-price, Recent, Recommend, Continuation, Expectation                              \\ \hline
    Quant. (Level 1)   & Soundness, Margins, Growth, Profitability, Growth-rate, Favorable, Deterioration, Financials, Issues, Unstable                        & EPS, Stability, Improvement, Profit, Earnings, Trend, Stagnation, Increase, Attractive, Stability \\ \hline
    Sector (Level 2)   & Momentum, Soundness, Financials, Margins, Profitability, Growth-rate, Indicators, ROE, Efficiency, Undervalued                      & Overall, Trend, EPS, Improvement, Operating-profit, Stagnation, Net-income, Stock-price, Trend, Decrease                        \\ \hline
    PM (Level 3)      & Momentum, Soundness, Financials, Profitability, Undervalued, Issues, Undervaluation, Margins, Overvaluation, Attractive & Sector, Trend, Improvement, Stagnation, EPS, Average, Stability, Decrease, Target, Downward    \\
    \bottomrule                          
    \end{tabular}
\end{table*}








\end{document}